\documentclass[12pt]{article}
\usepackage[letterpaper, portrait, margin=0.75in]{geometry}
\usepackage{amsmath,amssymb,amsthm,hyperref, graphicx}

\title{Morphological Error Detection in 3D Segmentations}
\author{
\small{\textbf{David Rolnick}$^1$\footnote{Correspondence should be addressed to: \texttt{drolnick@mit.edu}.}, \textbf{Yaron Meirovitch}$^1$, \textbf{Toufiq Parag}$^2$, \textbf{Hanspeter Pfister}$^2$}\\
\small{\textbf{Viren Jain}$^3$, \textbf{Jeff W.~Lichtman}$^2$, \textbf{Edward S.~Boyden}$^1$, \textbf{Nir Shavit}$^1$}\\
\small{$^1$Massachusetts Institute of Technology, $^2$Harvard University, $^3$Google, Inc.}
}
\date{}
\begin{document}

\maketitle

\begin{abstract}
Deep learning algorithms for connectomics rely upon localized classification, rather than overall morphology.  This leads to a high incidence of erroneously merged objects. Humans, by contrast, can easily detect such errors by acquiring intuition for the correct morphology of objects.  Biological neurons have complicated and variable shapes, which are challenging to learn, and merge errors take a multitude of different forms. We present an algorithm, MergeNet, that shows 3D ConvNets can, in fact, detect merge errors from high-level neuronal morphology.  MergeNet follows unsupervised training and operates across datasets.  We demonstrate the performance of MergeNet both on a variety of connectomics data and on a dataset created from merged MNIST images.
\end{abstract}

\section{Introduction}
The neural network of the brain remains a mystery, even as engineers have succeeded in building artificial neural networks that can solve a wide variety of problems.  Understanding the brain at a deeper level could significantly impact both biology and artificial intelligence \cite{bengio2015towards, helmstaedter2013connectomic, marblestone2016toward, fuzzy, takemura2013visual}.  Perhaps appropriately, artificial neural networks are now being used to map biological neural networks.  However, humans still outperform computer vision algorithms in segmenting brain tissue.  Deep learning has not yet attained the intuition that allows humans to recognize and trace the fine, intermingled branches of neurons.

The field of \emph{connectomics} aims to reconstruct three-dimensional networks of biological neurons from high-resolution microscope images.  Automated segmentation is a necessity due to the quantities of data involved.  In one recent study \cite{hildebrand2017whole}, the brain of a larval zebrafish was annotated by hand, requiring more than a year of human labor.  It is estimated that mapping a single human brain would require a zettabyte (one billion terabytes) of image data \cite{lichtman2014big}, clearly more than can be manually segmented.

State-of-the-art algorithms apply a convolutional neural network (ConvNet) to predict, for each voxel of an image, whether it is on the boundary (\emph{cell membrane}) of a neuron.  The predicted membranes are then filled in by subsequent algorithms \cite{jain2011learning}.  Such methods are prone both to \emph{split errors}, in which true objects are subdivided, and to \emph{merge errors}, in which objects are fused together.  The latter pose a particular challenge.  Neurons are highly variable, unpredictably sprouting thousands of branches, so their correct shapes cannot be catalogued.  Erroneously merged neurons are obvious to trained humans because they simply don't look right, but it has hitherto been impossible to make such determinations automatically.

We introduce a deep learning approach for detecting merge errors that leverages the morphological intuition of human annotators.  Instead of relying upon voxelwise membrane predictions or microscope images, we zoom out and capture as much context as possible.  Using only three-dimensional binary masks, our algorithm is able to learn to distinguish the shapes of plausible neurons from those that have been erroneously fused together.

We test our network, MergeNet, both on connectomics datasets and on an illustrative dataset derived from MNIST \cite{lecun1998mnist}.  The key contributions of this approach include:
\begin{itemize}
\item \textbf{Localization of merge errors.}
MergeNet is able to detect merge errors with high accuracy within a three-dimensional segmentation and to pinpoint their locations for correction (see Figures \ref{fig:heatmap} and \ref{fig:training_data}).

\item \textbf{Unsupervised training.}
The algorithm can be trained using any reasonably accurate segmentation, without the need for any additional annotation.  It is even able to correct errors within its own training data.
\begin{figure}
\centering
\begin{minipage}{0.48\textwidth}
  \centering
\includegraphics[width=1\linewidth]{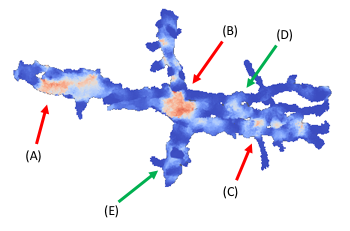}
\caption{\label{fig:heatmap}A probability map localizing merge errors, as predicted by MergeNet, for an object within the ECS dataset.  Orange indicates a high probability of merge error, blue the absence of error. Location (A) illustrates a merge between two neurons running in parallel, (B) a merge between three neurons simultaneously (the two parallel neurons, plus a third perpendicular to them), and (C) a merge between a large neuron segment and a small branch from another neuron.  MergeNet is able to learn that all of these diverse morphologies (and others not illustrated in this example) represent merge errors, but that locations such as (D) and (E) are normally occurring branch points within a single neuron.}
\end{minipage}%
\hspace{4pt}
\begin{minipage}{0.48\textwidth}
  \centering
\includegraphics[width=1\linewidth]{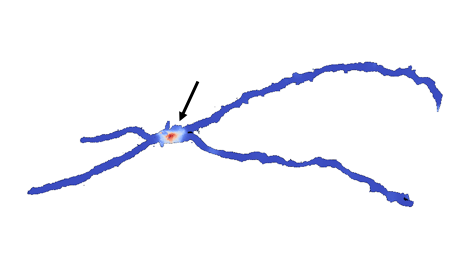}
\caption{\label{fig:training_data}A single, relatively simple merge error detected and localized by MergeNet. This object, within the Kasthuri dataset \cite{kasthuri2015saturated}, occurred \emph{within the training data} of the algorithm, but was not labeled as a merge error. MergeNet was nonetheless able to correct the label. This capability allows MergeNet to be trained on an uncertain segmentation, then used to correct errors within the same segmentation, without requiring any manual annotation.}
\vspace{70pt}
\end{minipage}
\end{figure}

\item \textbf{Generalizability and scalability across datasets.}
MergeNet can be applied irrespective of the segmentation algorithm or imaging method.  It can be trained on one dataset and run on another with high performance.  By downsampling volumetric data, our ConvNet is able to process three million voxels a second, faster than most membrane prediction systems.
\end{itemize}

\section{Related work}

There have been numerous recent advances in using neural networks to recognize general three-dimensional objects.  Methods include taking 2D projections of the input \cite{projections}, combined 2D-3D approaches \cite{chen2016combining, lee2015recursive}, and purely 3D networks \cite{maturana2015voxnet, wu20153d}. Accelerated implementation techniques for 3D networks have been introduced by Budden, et al.~\cite{budden2016deep} and Zlateski, Lee, and Seung~\cite{zlateski2017scalable}.

Within the field of connectomics, Maitin-Shepard et al.~\cite{maitin2016combinatorial} describe CELIS, a neural network approach for optimizing local features of a segmented image.  Januszewski et al.~\cite{januszewski2016flood} and Meirovitch et al.~\cite{meirovitch2016multi} present approaches for directly segmenting individual neurons from microscope images, without recourse to membrane prediction and agglomeration algorithms.  Deep learning techniques have likewise been used to detect synapses between neurons \cite{roncal2014vesicle,santurkar2017toward} and to localize voltage measurements in neural circuits \cite{apthorpe2016automatic} (progress towards a \emph{functional connectome}).  New forms of data are also being leveraged for connectomics \cite{peikon2017using, sumbul2016automated}, thanks to advances in biochemical engineering.

Many authors cite the frequent problems posed by merge errors (see e.g.~\cite{parag2015efficient}); however, almost no approaches have been proposed for detecting them automatically.  Meirovitch et al.~\cite{meirovitch2016multi} suggest a hard-coded heuristic to find ``X-junctions'', one variety of merge error, by analyzing graph theoretical representations of neurons as \emph{skeletons} (see also \cite{zhao2014automatic}).  Recent work including \cite{kipf2016semi, niepert2016learning} has considered the problem of deep learning on graphs, and Farhoodi, Ramkumar, and Kording \cite{konradgans} use Generative Adversarial Networks (GANs) to generate neuron skeletons.  However, such methods have not to date been brought to bear on connectomic reconstruction of neural circuits.

\section{Methods}
Our algorithm, MergeNet, operates on an image segmentation to correct errors within it.  Given an object within the proposed segmentation, MergeNet determines whether points chosen within the object are the location of erroneous merges.  If no such points exist, then the object is determined to be free from merge errors.

\textbf{Input and architecture.}\\
The input to our network is a three-dimensional \emph{window} of the object in question, representing a $51\times 51\times 51$ section of the object, centered at the chosen point.  (These dimensions are chosen as a tradeoff between enhancing speed and capturing more information, as we discuss further in sections \S\ref{sec:results} and \ref{sec:discussion}.) Crucially, the window is given as a \emph{binary mask}: that is, each voxel is 0 or 1 depending on whether it is assigned to the object.  MergeNet is not given data from the original image, inducing the network to learn general morphological features present in the binary mask.  The network follows a simple convolutional architecture, containing six convolutional layers with rectified linear unit (ReLU) activation, and three max-pooling layers, followed by a densely connected layer and softmax output.  The desired output is a 1-hot vector for the two classes  ``merge'' and ``no merge'', and is trained with cross-entropy loss.

\textbf{2D MergeNet.}\\
We also constructed a simpler 2D version of MergeNet to illustrate the identification of merge errors within two-dimensional images.  In this case, the input to the network is a square binary mask, which passes through four convolutional layers and two max-pooling layers. This network was trained to recognize merges between binarized digits from the MNIST dataset \cite{lecun1998mnist}.  Random digits were drawn from the training dataset and chained together, with merge errors given by pixels at the points of contact between neighboring digits.  Testing was performed on similar merges created from the testing dataset.  The size of the input window to the network was varied to compare accuracy across a variety of contextual scales.

\textbf{Downsampling.}\\
To apply MergeNet to connectomics data, we begin by downsampling all objects.  Segmentations of neural data are typically performed at very high resolution, approximately 5 nm.  The finest morphological details of neurons, however, are on the order of 100 nm.  Commonly, data is anisotropic, with resolution in the $z$ direction being significantly lower than that in $x$ and $y$. We tested MergeNet with downsampling ratios of $10 \times 10 \times 2$ and $25 \times 25 \times 5$ to compensate for anisotropy.  Downsampling an object is performed by a max-pooling procedure. That is, every voxel within the downsampled image represents the intersection of the object with a corresponding subvolume of the original image with dimensions e.g.~$25 \times 25 \times 5$.

\textbf{Training.}\\
The network was trained on artificially induced merge errors between objects within segmentations of neural tissue.  Merges consisted of identifying immediately adjacent objects and designating points of overlap as the locations of merge errors.  Negative examples consisted of windows centered at random points of objects within the segmentation.  Artificial merge errors are used owing to the impracticality of manually annotating data over large enough volumes to determine sufficient merge errors for training. As we demonstrate, such training suffices for effective detection of real merge errors and has numerous other advantages (detailed in section \S\ref{sec:discussion}).  Training was performed on various segmentations of the Kasthuri dataset \cite{kasthuri2015saturated}, and the algorithm was evaluated both on this dataset and on segmented objects of the ECS dataset, a $20\times 20\times 20$ micron cube of rat cortex data to which we were given access.

\textbf{Output.}\\
To run a trained instance of MergeNet on objects within a segmentation, it is not necessary to apply the network to every voxel since the predictions at nearby points may be interpolated.  Sample points are therefore taken within each downsampled object, and MergeNet is run on windows centered at these points. The real-valued predictions at sample points are then interpolated over the entire object to give a heatmap of probabilities for merge errors.  As the distribution of training examples is balanced between positive and negative examples, which is not true when the network is applied in practice, the output must be normalized or thresholded after the softmax layer.  We find it effective to classify as a merge error any voxel at which the prediction exceeds $0.9$.

\section{Results}
\label{sec:results}
\begin{figure}[htbp]
\begin{center}
\includegraphics*[scale=0.5]{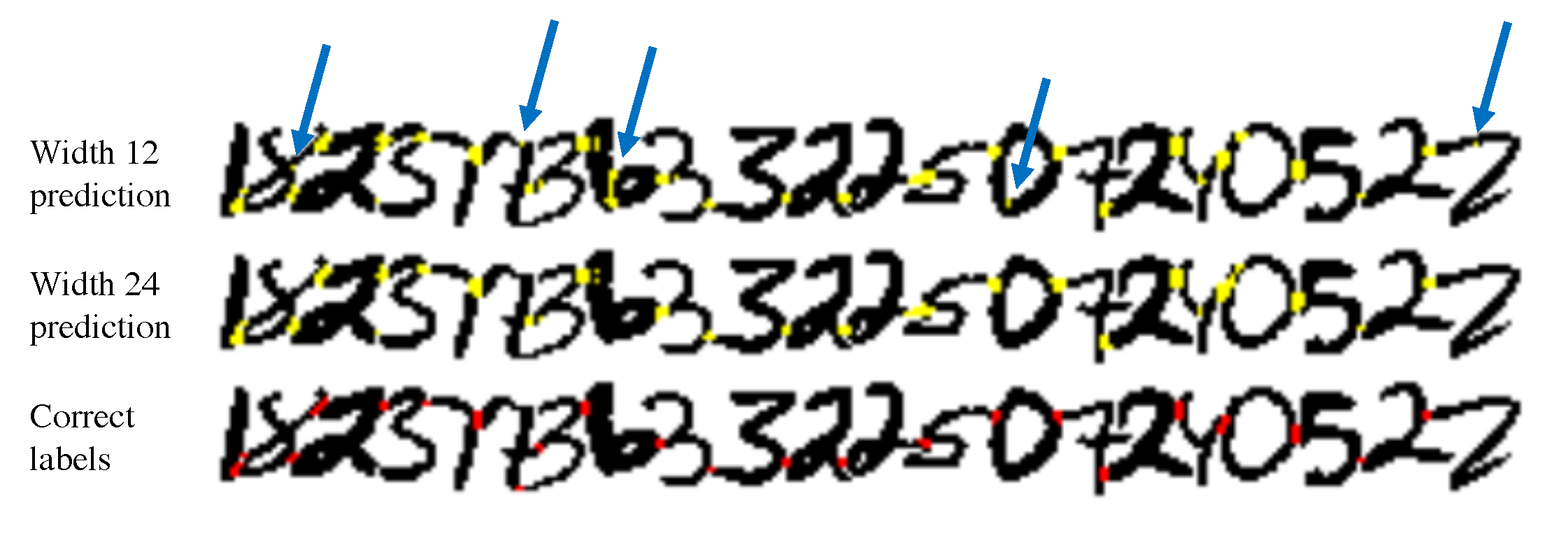}
\caption{Predictions of MergeNet on merged MNIST digits, shown on a sample of the test set. The top image shows predictions of the network with $12\times 12$ input windows, with predicted merges shown in yellow.  The middle image shows predictions with $24\times 24$ input windows.  The final image shows the actual merges, in red.  Note that both networks are quite accurate, but that for $12\times 12$ input, the algorithm makes several erroneous predictions of merge errors (shown with blue arrows), which are not made for $24\times 24$ input.  This illustrates how greater morphological context leads to qualitatively better predictions.  A quantitative assessment is shown in Figure \ref{fig:window_size}.}
\label{fig:mnist}
\end{center}
\end{figure}

We first consider the illustrative example of the merged MNIST dataset.  After training on five million $  $ examples within this dataset, we obtained a maximum pixelwise accuracy of ${\bf 96.8}$ \textbf{percent} on a test set constructed from held out digits and equally distributed between positive and negative examples.  This corresponds to almost perfect identification of individual merge error regions, as shown in Figure \ref{fig:mnist}.  (Ambiguous pixels on the edges of merge error regions were the most likely pixels to be misclassified.)

\begin{figure}[htbp]
\begin{center}
\includegraphics*[scale=0.5]{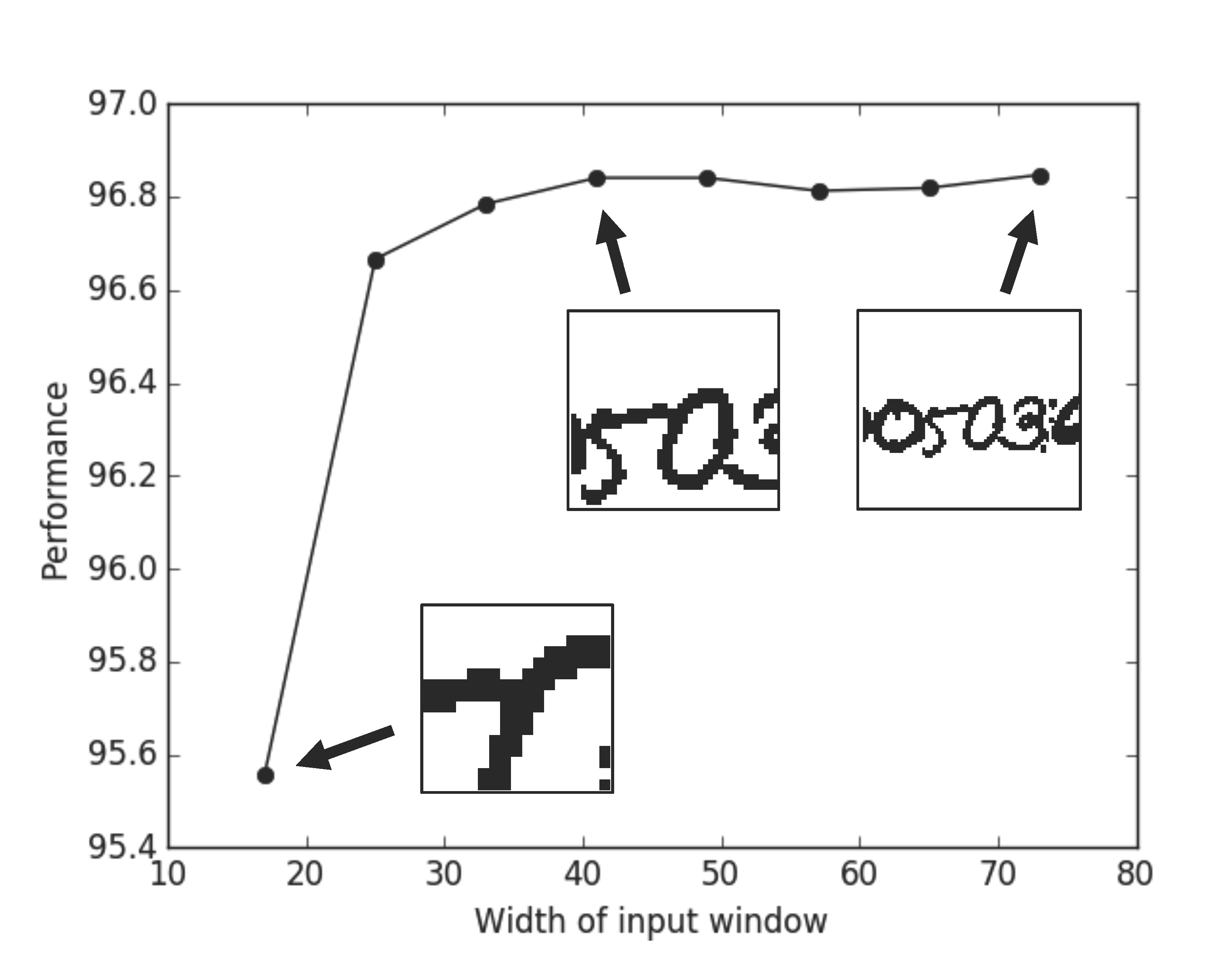}
\caption{Performance of MergeNet on merged MNIST digits, shown as a function of input window width (with example windows shown). Note that increasing window size increases performance, but only up to a size of 40-50 pixels. We see that performance plateaus at the point at which morphological context captures all of the information about the neighboring digits. For the case of neurons, which are larger and more complicated, morphological context does not plateau; however, there is a tradeoff between more context and greater speed, since the time required to run the 3D ConvNet depends strongly upon input size.}
\label{fig:window_size}
\end{center}
\end{figure}

\textbf{Accuracy increases with morphological information.}\\
In Figure \ref{fig:window_size}, we show the dependence of accuracy on the window size used.  Intuitively, a larger window gives the network more morphological context to work from, and plateaus in this case at approximately the dimensions of a pair of fused MNIST digits (40-50 pixels across), which represents the maximum scale at which morphology is useful to the network.  Figure \ref{fig:mnist} provides a qualitative comparison of performance between a smaller and a larger window size.  The smaller window size erroneously predicts merges within digits, while the large window size allows the network to recognize the shapes of these digits and identify only merge errors between digits.

There is, of course, a tradeoff between accuracy and the time required to train and run the network. Slowdown resulting from larger window size is considerable for three-dimensional ConvNets.  We have attempted to choose the parameters of MergeNet with this tradeoff in mind.

\textbf{MergeNet detects merge errors across datasets.}\\
We trained MergeNet on two segmentations of the Kasthuri dataset \cite{matveev2017multicore} and tested performance on artificially merged objects omitted from the training set. Segmentation A was relatively poor, and training on it yielded performance of only \textbf{77.3 percent}. Segmentation B was more accurate, yielding performance of \textbf{90.0 percent}. Training on both segmentations together yielded the best performance on both test sets, showing that MergeNet is able to leverage contextual information from one segmentation to improve performance on another.  We also note that after training on the low-quality Segmentation A, MergeNet was able to detect errors \emph{within its own training set}, as shown in Figure \ref{fig:training_data}.
\begin{table}[htbp]
\begin{tabular}{c|ccc}
& Training on Seg.~A & Training on Seg.~B & \textbf{Training on both} \\ \hline
Testing on Seg.~A & 77.3\% & 82.6\% & \textbf{84.5\%}\\
Testing on Seg.~B & 70.7\% & 90.0\% & \textbf{91.9\%} \\
\end{tabular}
\end{table}

MergeNet generalizes broadly across datasets as well as segmentations of the same dataset, and applies to both artificial and natural merge errors, though the latter are harder to quantify owing to the paucity of large-scale annotation. After training on the Kasthuri dataset, MergeNet was able to detect naturally occurring merge errors within segmentations of the ECS dataset, obtained from a state-of-the-art U-Net segmentation algorithm \cite{ronneberger2015u}. Example output is shown in Figures \ref{fig:heatmap} and \ref{fig:membranes}.

\section{Discussion}
\label{sec:discussion}

We will now consider the capabilities of the MergeNet algorithm and discuss opportunities that it offers within the field of connectomics.

\textbf{Detection and localization of merge errors.}\\
MergeNet is a powerful tool for detecting and pinpointing merge errors.  Once a merge location has been flagged with high spatial precision, other algorithms can be used to create a more accurate local segmentation, thereby correcting any errors that occured.  Flood-filling networks \cite{januszewski2016flood} and MaskExtend \cite{meirovitch2016multi} are two examples of algorithms that have high accuracy, but are extremely time-consuming to run over large volumes, making them ideally suited to segment at the merge locations flagged by MergeNet.  Alternatively, the agglomeration algorithm NeuroProof \cite{parag2015context}, used in transforming membrane probabilities to segmentations, can be tuned to be more or less sensitive to merge errors.  A more merge-sensitive setting could be applied at those locations flagged by our algorithm.

If the thresholding step is omitted, then the output of MergeNet may be thought of as a probability distribution of merge errors over objects within a segmentation.  This distribution may be treated as a Bayesian prior and updated if other information is available; multiple proofreading algorithms can work together.  Thus, for example, synapse detection algorithms \cite{roncal2014vesicle,santurkar2017toward} may provide additional evidence for a merge error if synapses of two kinds are found on the same segmented object, but are normally found on different types of neurons.  In such a scenario, the probability at the relevant location would be increased from its value computed by MergeNet, according to the confidence of the synapse detection algorithm.  We envision MergeNet being the first step towards fully automated proofreading of connectomics data, which will become increasingly necessary as such data is processed at ever greater scale.

\textbf{Unsupervised training.}\\
MergeNet is trained on merge errors created by fusing adjacent objects within a segmentation.  This allows training to proceed without any direct human annotation.  In testing MergeNet, we performed training on several automatically generated segmentations of EM data and obtained good results, even though the training data was not free of merge errors and other mistakes.  It is highly advantageous to eliminate the need for further data annotation, since this is the step in connectomics that has traditionally consumed by far the most human effort.   The ability to run on any (reasonably accurate) segmentation also means that MergeNet can be trained on far larger datasets than those for which manual annotation could reasonably be obtained.  Automated segmentations already exist for volumes of neural tissue as large as 232,000 cubic microns \cite{plaza2016large}.

\textbf{Comparison of segmentations.}\\
Automatic detection of merge errors allows us to compare the performance of alternative segmentation algorithms \emph{in the absence of ground truth annotation}.  We ran MergeNet with the same parameter settings on two segmentations within the ECS dataset, after training on other data.  These alternative segmentations were produced by two different versions of a state-of-the-art U-Net segmentation algorithm \cite{ronneberger2015u}.  For the simpler algorithm, 33 of the 300 largest objects within the segmentation (those most likely to have merge errors) were flagged as unlikely, while, for the more advanced algorithm, only 15 of the largest 300 objects were flagged.  This indicates that the latter pipeline produces more plausible objects, making fewer merge errors.  The size of the objects was comparable in each case, so there is no indication that this improvement came with a greater propensity for erroneous splits within single objects.  Thus, MergeNet was able to perform a fully automatic comparison of two segmentation algorithms and confirm that one outperforms the other.

\textbf{Correction of the training set.}\\
We have already observed that MergeNet can be trained on any reasonably correct segmentation.  In fact, it is possible to leverage artificial merge errors within the training set to detect real merge errors that may occur there (as shown in Figure \ref{fig:training_data}).  This is remarkable, since the predictions MergeNet makes in this case should conflict with the labels it was itself trained on - namely, when objects from the training set that have not been artificially merged are nonetheless the result of real merge errors.  Our observations align with results showing that neural networks are capable of learning from data even when the labels given are unreliable \cite{sukhbaatar2014training}.

\textbf{Independence from lower-order errors.}\\
Since MergeNet makes use of the global morphology of neurons, it is not reliant on earlier stages of the connectomics pipeline, such as microscope images or membrane predictions.  Thus, it is able to correct errors that arise at early stages of the pipeline, including those at the experimental stage.  EM images are prone to various catastrophic errors; most notably, individual tissue slices can tear before imaging, leading to distortion or gaps in the predicted membranes.  Algorithms that stitch together adjacent microscope images also sometimes fail, leading to pieces of neurons in one image being erroneously aligned with pieces from the neighboring image.  Typically, such errors are propagated or magnified by later stages of the connectomics pipeline, since algorithms such as watershed and NeuroProof \cite{parag2015context} assume that their input is mostly true.  By contrast, MergeNet can look at the broader picture and use ``common sense'' as would a human proofreader.

\begin{figure}
\centering
\begin{minipage}{0.48\textwidth}
  \centering
\includegraphics[width=0.7\linewidth]{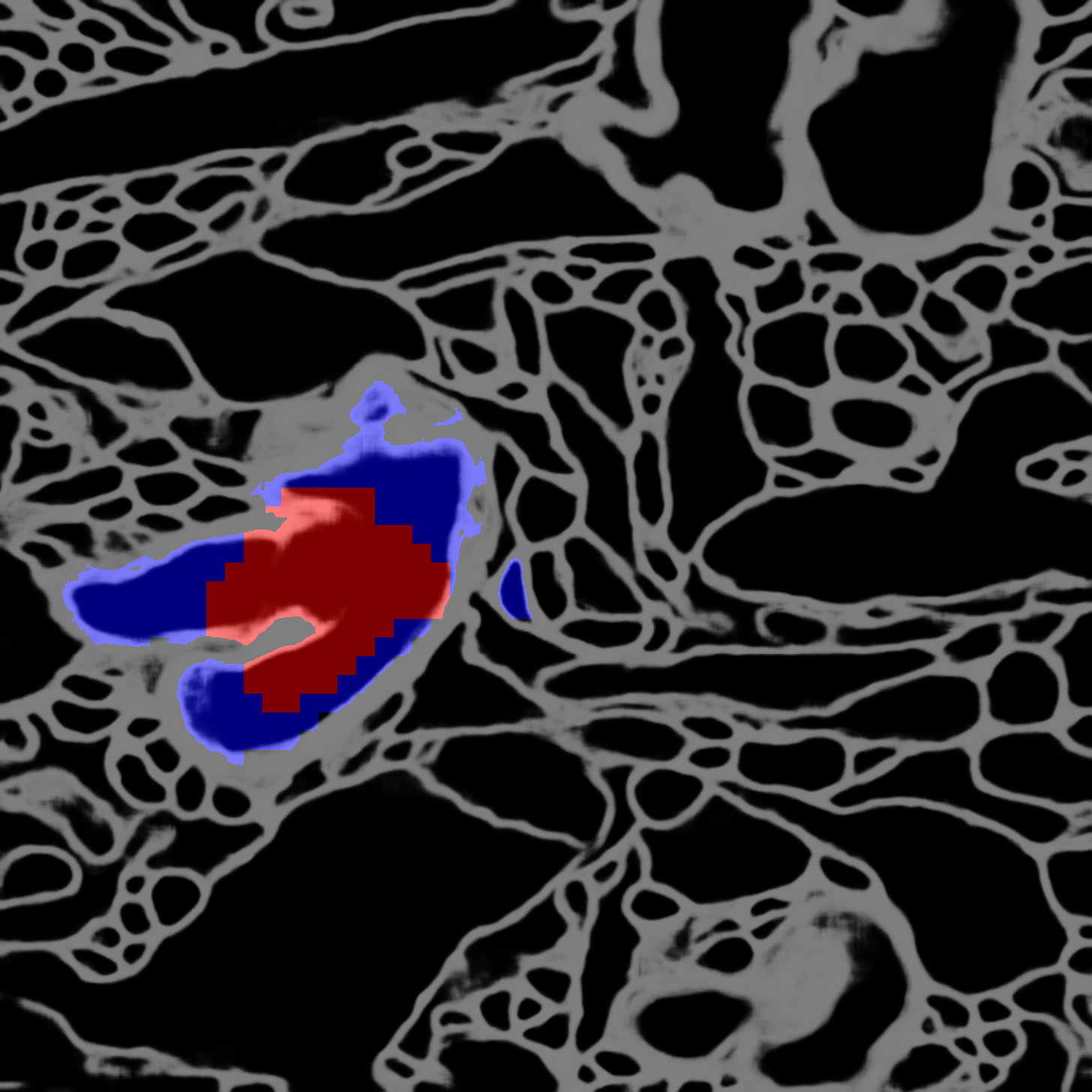}
\caption{\label{fig:membranes}The output of MergeNet at a merge error, superimposed over the erroneously predicted membranes that led to the merge error.  MergeNet output at individual pixels has been thresholded above $0.9$, with red denoting predicted merge error and blue the absence of error.  Observe that the predicted membranes have a wide gap at the region MergeNet has flagged; this gap is incorrect and the membrane should extend between the two objects, separating them. Note that MergeNet used only three-dimensional morphological information to detect this error, and did not make use of the (erroneous) membrane predictions that are shown, or the underlying microscope images.}
\end{minipage}%
\hspace{4pt}
\begin{minipage}{0.48\textwidth}
  \centering
\includegraphics[width=0.7\linewidth]{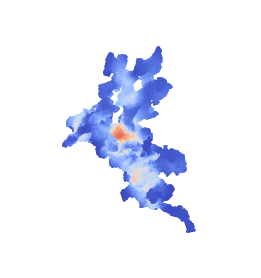}
\caption{\label{fig:glia}Glial cell, flagged as a merge error by MergeNet. While glia are not merge errors, they are also not neurons and did not occur in the training set for MergeNet. As the algorithm recognizes, the morphology of glia is markedly different from that of neurons. Specifically training MergeNet to recognize glia could be useful in segmenting these cells, which occur along with neurons in brain tissue.}
\vspace{100pt}
\end{minipage}
\end{figure}

The output of a membrane-detection algorithm also can induce errors in object morphology.   Common sources of error at this stage are ambiguously stained tissue slices and intracellular membranes, such as those from mitochondria, which can be confused with the external cell membrane.  Figure \ref{fig:membranes} shows an error in predictions from the U-Net algorithm, where a large gap in the predicted membrane has allowed two objects to be fused together into one (shown in blue).  By utilizing the overall 3D context, MergeNet is able to detect and localize the error (shown in red).

\textbf{Generalizability to different datasets.}\\
One of the challenges of traditional connectomics algorithms is that there are numerous different imaging techniques, which can each be applied to the nervous systems of various organisms, and in some cases also to structurally distinct regions of the nervous system within a single organism.  For networks used in connectomics for image segmentation, it is often necessary to obtain ground truth annotations on each new dataset, which consumes considerable time and effort.

MergeNet, by contrast, is highly transferrable between datasets.  Not only can the algorithm be trained on an unverified segmentation and can correct it, but it can also be trained on one dataset and then run on a segmentation of a different dataset, without any retraining. Figures \ref{fig:heatmap} and \ref{fig:membranes} show images obtained by training on a segmentation of the Kasthuri dataset \cite{kasthuri2015saturated} and then running on the ECS dataset.

\textbf{Applicability to anisotropic data.}\\
The microscope data underlying connectomics segmentation is often anisotropic, where the particular dimensions in the $x,y,z$ directions depend upon the particular imaging procedure used.  For example, the Kasthuri dataset has resolution of $6\times 6\times 30$ nanometers, while our ECS dataset is even more anisotropic, with resolution of $4\times 4\times 30$ nanometers. Some imaging technologies do yield isotropic data, such as expansion microscopy (ExM) \cite{chen2015expansion} and focused ion beam scanning electron microscopy (FIB-SEM) \cite{knott2008serial}.  Various techniques have been proposed to work with anisotropic data, including 2D ConvNets feeding into 3D ConvNets \cite{lee2015recursive} and a combination of convolutional and recurrent networks \cite{chen2016combining}.

MergeNet cancels the effect of anisotropy, as necessary, by downsampling differentially along the $x,y,z$ directions.  Thus, the network is able to transform any segmentation into one in which morphology is approximately isotropic, making learning much easier.  We also anticipate that it may be possible to train on data from one imaging modality, then to apply a different downsampling ratio to run on data with different anisotropy.  For example, MergeNet could be trained on an EM segmentation, then run on an ExM segmentation.

\textbf{Scalability.}\\
The MergeNet algorithm is designed to be scalable, so that it can be used to proofread segmentations of extremely large datasets.  The network is applied only once objects have been downsampled by a large factor in each dimension, and is applied then only to sampled points within the downsampled object.  These two reductions in cost allow the 3D ConvNet to be run at scale, even though 3D kernels are slower to implement than 2D kernels.

We tested the speed of MergeNet on an object with 36,874,153 original voxels, downsampled to 18,777 voxels, from which we sampled 1,024, allowing us to generate a dense probability map across the entire object.  The ConvNet ran in 11.3 seconds on an Nvidia Tesla K20m GPU. This corresponds to a speed of over three million voxels per second within the original image.  Thus, the network could be applied to a volume of 1 billion voxels in a minute using five GPUs.  By comparison, the fastest membrane-prediction algorithm can process 1 billion voxels within 2 minutes on a 72-core machine \cite{matveev2017multicore}, demonstrating that our algorithm can be integrated into a scalable connectomics pipeline.  Note that our experiments were performed using TensorFlow \cite{abadi2016tensorflow}; we have not attempted to optimize time for training or running the network, though recent work indicates that significant further speedup may be possible \cite{budden2016deep, zlateski2017scalable}.

\textbf{Detection of non-neuronal objects.}\\
While MergeNet is trained only on merge errors, it also seems to be able to detect non-neuronal objects, as a byproduct of learning plausible shapes for neurons.  In particular, we observe that MergeNet often detects glia (nonneuronal cells that occur in neural tissue), the morphologies of which are distinctively different from those of neurons.  Figure \ref{fig:glia} shows an example of a glial object from the ECS dataset; notice that MergeNet finds the morphology implausible, even though it has been trained on neither positive nor negative examples of glia.  Quantifying the accuracy of glia detection is challenging, however, since little ground truth has been annotated for this task, and most connectomics algorithms are unable to distinguish glia.

Finally, let us consider when (and why) MergeNet succeeds and fails on different inputs. The algorithm does not simply label all branch points within a neuron as merge errors, or else it would be effectively useless.  However, the network can be confused by examples such as two branches that diverge from a main segment at approximately the same point, resembling the cross of two distinct objects. MergeNet also misses some merge errors.  For example, when two neuronal segments run closely in parallel, there may at some points along the boundary be no morphological clues that two objects are present.  It is worth noting, however, that parallel neuronal segments can in fact be detected by MergeNet, as shown at point (A) of Figure \ref{fig:heatmap}.

\section{Conclusion}
Though merge errors occur universally in automated segmentations of neural tissue, they have never been addressed in generality, as they are difficult to detect using existing connectomics methods. We have shown that a 3D ConvNet can proofread a segmented image for merge errors by ``zooming out'', ignoring the image itself, and instead leveraging the general morphological characteristics of neurons.  We have demonstrated that our algorithm, MergeNet, is able to generalize without retraining to detect errors within a range of segmentations and across a range of datasets.  Relying solely upon unsupervised training, it can nonetheless detect errors within its own training set.  Our algorithm enables automatic comparison of segmentation methods, and can be integrated at scale into existing pipelines without the requirement of additional annotation, opening up the possibility of fully automated error detection and correction within neural circuits.

While MergeNet can detect and localize merge errors, it cannot, by itself, correct them.  One could conceive a variation on the MergeNet algorithm that is used to mark the exact boundary of a merge error, allowing a cut to be made automatically along the boundary so as to correct the merge without additional effort.  However, in practice this is a much more challenging task.  Often it is impossible to determine the exact division of objects at a merge error purely from morphology.  For instance, when two largely parallel objects touch, it may not be evident which is the continuation of which past the point of contact, even if the erroneous merge itself is obvious.  Likewise, some merge errors consist of three or more objects that have been confused in some complex way, e.g.~by virtue of poor image quality at that location.  In such cases, any merge-correction algorithm must have recourse to the underlying microscope images or membrane probabilities, rather than relying purely upon morphological cues.

Deep learning approaches leveraging morphology have the potential to transform biological image analysis. It may, for instance, become possible to classify types of neurons automatically, or to identify anomalies such as cancer cells.  We anticipate a growth in such algorithms as the scale of biological data grows and as progress in connectomics leads to a deeper understanding of the brain.

\section{Acknowledgments}
We are grateful for support from the National Science
Foundation (NSF) under grants IIS-1447786, CCF-
1563880, and 1122374 and from the Intelligence Advanced Research Projects
Activity (IARPA) under grant 138076-5093555.  We would like to thank Tom Dean, Peter Li, Art Pope, Jeremy Maitin-Shepard, Adam Marblestone, and Hayk Saribekyan for helpful discussions and contributions.

\small{
\bibliographystyle{plain}
\bibliography{references}

\begin{thebibliography}{10}

\bibitem{abadi2016tensorflow}
Mart{\'\i}n Abadi, Ashish Agarwal, Paul Barham, Eugene Brevdo, Zhifeng Chen,
  Craig Citro, Greg~S Corrado, Andy Davis, Jeffrey Dean, Matthieu Devin, et~al.
\newblock Tensorflow: Large-scale machine learning on heterogeneous distributed
  systems.
\newblock {\em arXiv preprint arXiv:1603.04467}, 2016.

\bibitem{apthorpe2016automatic}
Noah Apthorpe, Alexander Riordan, Robert Aguilar, Jan Homann, Yi~Gu, David
  Tank, and H~Sebastian Seung.
\newblock Automatic neuron detection in calcium imaging data using
  convolutional networks.
\newblock In {\em Advances In Neural Information Processing Systems (NIPS)},
  pages 3270--3278, 2016.

\bibitem{bengio2015towards}
Yoshua Bengio, Dong-Hyun Lee, Jorg Bornschein, Thomas Mesnard, and Zhouhan Lin.
\newblock Towards biologically plausible deep learning.
\newblock {\em Preprint arXiv:1502.04156}, 2015.

\bibitem{budden2016deep}
David Budden, Alexander Matveev, Shibani Santurkar, Shraman~Ray Chaudhuri, and
  Nir Shavit.
\newblock Deep tensor convolution on multicores.
\newblock {\em arXiv preprint arXiv:1611.06565}, 2016.

\bibitem{chen2015expansion}
Fei Chen, Paul~W Tillberg, and Edward~S Boyden.
\newblock Expansion microscopy.
\newblock {\em Science}, 347(6221):543--548, 2015.

\bibitem{chen2016combining}
Jianxu Chen, Lin Yang, Yizhe Zhang, Mark Alber, and Danny~Z Chen.
\newblock Combining fully convolutional and recurrent neural networks for 3d
  biomedical image segmentation.
\newblock In {\em Advances in Neural Information Processing Systems (NIPS)},
  pages 3036--3044, 2016.

\bibitem{konradgans}
Roozbeh Farhoodi, Pavan Ramkumar, and Konrad Kording.
\newblock Deep learning approach towards generating neuronal morphology.
\newblock Cosyne Abstracts, Salt Lake City USA, 2017.

\bibitem{helmstaedter2013connectomic}
Moritz Helmstaedter, Kevin~L Briggman, Srinivas~C Turaga, Viren Jain,
  H~Sebastian Seung, and Winfried Denk.
\newblock Connectomic reconstruction of the inner plexiform layer in the mouse
  retina.
\newblock {\em Nature}, 500(7461):168--174, 2013.

\bibitem{hildebrand2017whole}
David Grant~Colburn Hildebrand, Marcelo Cicconet, Russel~Miguel Torres, Woohyuk
  Choi, Tran~Minh Quan, Jungmin Moon, Arthur~Willis Wetzel, Andrew~Scott
  Champion, Brett~Jesse Graham, Owen Randlett, et~al.
\newblock Whole-brain serial-section electron microscopy in larval zebrafish.
\newblock {\em Nature}, 2017.

\bibitem{jain2011learning}
Viren Jain, Srinivas~C Turaga, K~Briggman, Moritz~N Helmstaedter, Winfried
  Denk, and H~Sebastian Seung.
\newblock Learning to agglomerate superpixel hierarchies.
\newblock In {\em Advances in Neural Information Processing Systems (NIPS)},
  pages 648--656, 2011.

\bibitem{januszewski2016flood}
Micha{\l} Januszewski, Jeremy Maitin-Shepard, Peter Li, J{\"o}rgen Kornfeld,
  Winfried Denk, and Viren Jain.
\newblock Flood-filling networks.
\newblock {\em arXiv preprint arXiv:1611.00421}, 2016.

\bibitem{kasthuri2015saturated}
Narayanan Kasthuri, Kenneth~Jeffrey Hayworth, Daniel~Raimund Berger,
  Richard~Lee Schalek, Jos{\'e}~Angel Conchello, Seymour Knowles-Barley, Dongil
  Lee, Amelio V{\'a}zquez-Reina, Verena Kaynig, Thouis~Raymond Jones, et~al.
\newblock Saturated reconstruction of a volume of neocortex.
\newblock {\em Cell}, 162(3):648--661, 2015.

\bibitem{kipf2016semi}
Thomas~N Kipf and Max Welling.
\newblock Semi-supervised classification with graph convolutional networks.
\newblock {\em arXiv preprint arXiv:1609.02907}, 2016.

\bibitem{knott2008serial}
Graham Knott, Herschel Marchman, David Wall, and Ben Lich.
\newblock Serial section scanning electron microscopy of adult brain tissue
  using focused ion beam milling.
\newblock {\em Journal of Neuroscience}, 28(12):2959--2964, 2008.

\bibitem{lecun1998mnist}
Yann LeCun, Corinna Cortes, and Christopher~JC Burges.
\newblock The {MNIST} database of handwritten digits, 1998.

\bibitem{lee2015recursive}
Kisuk Lee, Aleksandar Zlateski, Vishwanathan Ashwin, and H~Sebastian Seung.
\newblock Recursive training of 2d-3d convolutional networks for neuronal
  boundary prediction.
\newblock In {\em Advances in Neural Information Processing Systems (NIPS)},
  pages 3573--3581, 2015.

\bibitem{lichtman2014big}
Jeff~W Lichtman, Hanspeter Pfister, and Nir Shavit.
\newblock The big data challenges of connectomics.
\newblock {\em Nature neuroscience}, 17(11):1448--1454, 2014.

\bibitem{maitin2016combinatorial}
Jeremy~B Maitin-Shepard, Viren Jain, Michal Januszewski, Peter Li, and Pieter
  Abbeel.
\newblock Combinatorial energy learning for image segmentation.
\newblock In {\em Advances in Neural Information Processing Systems (NIPS)},
  pages 1966--1974, 2016.

\bibitem{marblestone2016toward}
Adam~H Marblestone, Greg Wayne, and Konrad~P Kording.
\newblock Toward an integration of deep learning and neuroscience.
\newblock {\em Frontiers in Computational Neuroscience}, 10, 2016.

\bibitem{maturana2015voxnet}
Daniel Maturana and Sebastian Scherer.
\newblock Voxnet: A 3d convolutional neural network for real-time object
  recognition.
\newblock In {\em Intelligent Robots and Systems (IROS), 2015 IEEE/RSJ
  International Conference on}, pages 922--928. IEEE, 2015.

\bibitem{matveev2017multicore}
Alexander Matveev, Yaron Meirovitch, Hayk Saribekyan, Wiktor Jakubiuk, Tim
  Kaler, Gergely Odor, David Budden, Aleksandar Zlateski, and Nir Shavit.
\newblock A multicore path to connectomics-on-demand.
\newblock In {\em Proceedings of the 22nd ACM SIGPLAN Symposium on Principles
  and Practice of Parallel Programming}, pages 267--281. ACM, 2017.

\bibitem{meirovitch2016multi}
Yaron Meirovitch, Alexander Matveev, Hayk Saribekyan, David Budden, David
  Rolnick, Gergely Odor, Seymour Knowles-Barley Jones, Raymond Thouis,
  Hanspeter Pfister, Jeff~William Lichtman, and Nir Shavit.
\newblock A multi-pass approach to large-scale connectomics.
\newblock {\em arXiv preprint arXiv:1612.02120}, 2016.

\bibitem{fuzzy}
Josh~Lyskowski Morgan, Daniel~Raimund Berger, Arthur~Willis Wetzel, and
  Jeff~William Lichtman.
\newblock The fuzzy logic of network connectivity in mouse visual thalamus.
\newblock {\em Cell}, 165(1):192--206, 2016.

\bibitem{niepert2016learning}
Mathias Niepert, Mohamed Ahmed, and Konstantin Kutzkov.
\newblock Learning convolutional neural networks for graphs.
\newblock In {\em Proceedings of the 33rd annual international conference on
  machine learning. ACM}, 2016.

\bibitem{parag2015context}
Toufiq Parag, Anirban Chakraborty, Stephen Plaza, and Louis Scheffer.
\newblock A context-aware delayed agglomeration framework for electron
  microscopy segmentation.
\newblock {\em PloS one}, 10(5):e0125825, 2015.

\bibitem{parag2015efficient}
Toufiq Parag, Dan~C Ciresan, and Alessandro Giusti.
\newblock Efficient classifier training to minimize false merges in electron
  microscopy segmentation.
\newblock In {\em Proceedings of the IEEE International Conference on Computer
  Vision (ICCV)}, pages 657--665, 2015.

\bibitem{peikon2017using}
Ian~D Peikon, Justus~M Kebschull, Vasily~V Vagin, Diana~I Ravens, Yu-Chi Sun,
  Eric Brouzes, Ivan~R Corr{\^e}a, Dario Bressan, and Anthony Zador.
\newblock Using high-throughput barcode sequencing to efficiently map
  connectomes.
\newblock {\em bioRxiv}, page 099093, 2017.

\bibitem{plaza2016large}
Stephen~M Plaza and Stuart~E Berg.
\newblock Large-scale electron microscopy image segmentation in {S}park.
\newblock {\em arXiv preprint arXiv:1604.00385}, 2016.

\bibitem{roncal2014vesicle}
William~Gray Roncal, Michael Pekala, Verena Kaynig-Fittkau, Dean~M Kleissas,
  Joshua~T Vogelstein, Hanspeter Pfister, Randal Burns, R~Jacob Vogelstein,
  Mark~A Chevillet, and Gregory~D Hager.
\newblock Vesicle: Volumetric evaluation of synaptic interfaces using computer
  vision at large scale.
\newblock {\em arXiv preprint arXiv:1403.3724}, 2014.

\bibitem{ronneberger2015u}
Olaf Ronneberger, Philipp Fischer, and Thomas Brox.
\newblock U-net: Convolutional networks for biomedical image segmentation.
\newblock In {\em International Conference on Medical Image Computing and
  Computer-Assisted Intervention}, pages 234--241. Springer, 2015.

\bibitem{santurkar2017toward}
Shibani Santurkar, David Budden, Alexander Matveev, Heather Berlin, Hayk
  Saribekyan, Yaron Meirovitch, and Nir Shavit.
\newblock Toward streaming synapse detection with compositional convnets.
\newblock {\em arXiv preprint arXiv:1702.07386}, 2017.

\bibitem{projections}
Hang Su, Subhransu Maji, Evangelos Kalogerakis, and Erik Learned-Miller.
\newblock Multi-view convolutional neural networks for 3d shape recognition.
\newblock In {\em Proceedings of the IEEE International Conference on Computer
  Vision (ICCV)}, pages 945--953, 2015.

\bibitem{sukhbaatar2014training}
Sainbayar Sukhbaatar, Joan Bruna, Manohar Paluri, Lubomir Bourdev, and Rob
  Fergus.
\newblock Training convolutional networks with noisy labels.
\newblock {\em arXiv preprint arXiv:1406.2080}, 2014.

\bibitem{sumbul2016automated}
Uygar S{\"u}mb{\"u}l, Douglas Roossien, Dawen Cai, Fei Chen, Nicholas Barry,
  John~P Cunningham, Edward Boyden, and Liam Paninski.
\newblock Automated scalable segmentation of neurons from multispectral images.
\newblock In {\em Advances in Neural Information Processing Systems (NIPS)},
  pages 1912--1920, 2016.

\bibitem{takemura2013visual}
Shin-ya Takemura, Arjun Bharioke, Zhiyuan Lu, Aljoscha Nern, Shiv Vitaladevuni,
  Patricia~K Rivlin, William~T Katz, Donald~J Olbris, Stephen~M Plaza, Philip
  Winston, et~al.
\newblock A visual motion detection circuit suggested by {D}rosophila
  connectomics.
\newblock {\em Nature}, 500(7461):175--181, 2013.

\bibitem{wu20153d}
Zhirong Wu, Shuran Song, Aditya Khosla, Fisher Yu, Linguang Zhang, Xiaoou Tang,
  and Jianxiong Xiao.
\newblock 3d shapenets: A deep representation for volumetric shapes.
\newblock In {\em Proceedings of the IEEE Conference on Computer Vision and
  Pattern Recognition}, pages 1912--1920, 2015.

\bibitem{zhao2014automatic}
Ting Zhao and Stephen~M Plaza.
\newblock Automatic neuron type identification by neurite localization in the
  {D}rosophila medulla.
\newblock {\em arXiv preprint arXiv:1409.1892}, 2014.

\bibitem{zlateski2017scalable}
Aleksandar Zlateski, Kisuk Lee, and H~Sebastian Seung.
\newblock Scalable training of 3d convolutional networks on multi-and
  many-cores.
\newblock {\em Journal of Parallel and Distributed Computing}, 2017.

\end{thebibliography}
}
\end{document}